\documentclass[letterpaper]{article} 
\usepackage{aaai2026}  
\usepackage{times}  
\usepackage{helvet}  
\usepackage{courier}  
\usepackage[hyphens]{url}  
\usepackage{graphicx} 
\usepackage{natbib}  
\usepackage{caption} 
\usepackage{algorithm}
\usepackage{algorithmic}

\usepackage{amsbsy}
\usepackage{bm}
\usepackage{amsmath}
\usepackage{threeparttable}
\usepackage{multirow}
\usepackage{booktabs}
\usepackage{amsthm}
\usepackage{amssymb} 
\usepackage[table]{xcolor}

\usepackage{times}
\usepackage{helvet}
\usepackage{courier}
\usepackage{xcolor}



\usepackage{newfloat}
\usepackage{listings}
\DeclareCaptionStyle{ruled}{labelfont=normalfont,labelsep=colon,strut=off} 
\floatstyle{ruled}
\newfloat{listing}{tb}{lst}{}
\floatname{listing}{Listing}
\title{Balancing Multimodal Domain Generalization via Gradient \\ Modulation and Projection}

\author{
    Hongzhao Li\textsuperscript{\rm 1},
    Guohao Shen\textsuperscript{\rm 1},
    Shupan Li\textsuperscript{\rm 1,2,3}\thanks{Corresponding authors.},
    Mingliang Xu\textsuperscript{\rm 1,2,3,4}\footnotemark[1],
    Muhammad Haris Khan\textsuperscript{\rm 5}
}

\affiliations{
    \textsuperscript{\rm 1}School of Computer and Artificial Intelligence, Zhengzhou University\\
    \textsuperscript{\rm 2}Engineering Research Center of Intelligent Swarm Systems, Ministry of Education\\
    \textsuperscript{\rm 3}National Supercomputing Center in Zhengzhou\\
    \textsuperscript{\rm 4}Shandong Bosuan Zhixin Information Technology Co., Ltd\\
    \textsuperscript{\rm 5}Mohamed Bin Zayed University of Artificial Intelligence\\
    \{lihongzhao, sgh\_2024\}@gs.zzu.edu.cn, \{iespli, iexumingliang\}@zzu.edu.cn, muhammad.haris@mbzuai.ac.ae
}

\usepackage{bibentry}

\begin{document}

\maketitle

\begin{abstract}
Multimodal Domain Generalization (MMDG) leverages the complementary strengths of multiple modalities to enhance model generalization on unseen domains. A central challenge in multimodal learning is optimization imbalance, where modalities converge at different speeds during training. This imbalance leads to unequal gradient contributions, allowing some modalities to dominate the learning process while others lag behind. Existing balancing strategies typically regulate each modality’s gradient contribution based on its classification performance on the source domain to alleviate this issue. However, relying solely on source-domain accuracy neglects a key insight in MMDG: modalities that excel on the source domain may generalize poorly to unseen domains, limiting cross-domain gains. To overcome this limitation, we propose Gradient Modulation Projection (GMP), a unified strategy that promotes balanced optimization in MMDG. GMP first decouples gradients associated with classification and domain-invariance objectives. It then modulates each modality’s gradient based on semantic and domain confidence. Moreover, GMP dynamically adjusts gradient projections by tracking the relative strength of each task, mitigating conflicts between classification and domain-invariant learning within modality-specific encoders. Extensive experiments demonstrate that GMP achieves state-of-the-art performance and integrates flexibly with diverse MMDG methods, significantly improving generalization across multiple benchmarks.
\end{abstract}


\begin{figure}[t]
    \flushleft
    \includegraphics[width=0.9\linewidth]{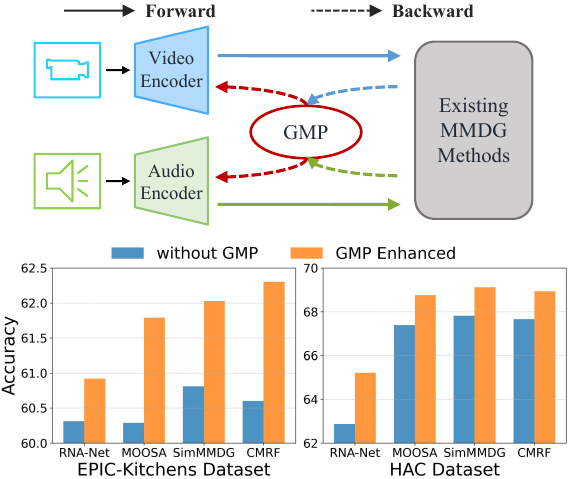}
    \caption{Performance of state of the art MMDG methods, RNA-Net \cite{planamente2024relative}, MOOSA \cite{dong2024moosa}, SimMMDG \cite{dong2023simmmdg}, and CMRF \cite{fan2024cross}, evaluated with and without our proposed GMP strategy.}
    \label{fig:teaser}
\end{figure}
\section{Introduction}
Multimodal Domain Generalization (MMDG) is an emerging research area focused on training models that generalize to unseen domains by leveraging multiple modalities, such as video and audio \cite{dong2025mmdasurvey,dong2025adapting}. Unlike traditional unimodal approaches \cite{munir2023domain,khan2024improving,galappaththige2024towards,galappaththige2024generalizing,li2026multimodaldomaingeneralizationlabels}, MMDG exploits the complementary strengths of different modalities to boost both robustness and generalization. This advantage is especially important in real-world applications like cross-environment action recognition, audiovisual event detection, and multimodal surveillance, where test data often originate from novel domains, such as different environments, devices, or conditions \cite{zhou2022domain}. In such cases, effective generalization hinges on learning domain invariant features without compromising discriminative power \cite{wang2022generalizing}.

A common challenge in multimodal learning is the optimization imbalance between modalities \cite{wei2024diagnosing}. During training, different modalities often converge at varying rates, leading to uneven gradient updates. When one modality dominates, it can hinder learning in others, disrupting balanced training dynamics. This imbalance reduces the effectiveness of multimodal learning, as the model becomes overly dependent on the dominant modality while underutilizing the rest \cite{peng2022balanced}. In MMDG, this problem is especially evident. As shown in Table \ref{tab:unimodal}, the performance of each unimodal branch in a typical video-audio MMDG model is significantly worse than when trained independently. This indicates that current MMDG training strategies fail to capitalize on the strengths of each modality, limiting the model’s ability to generalize across unseen domains.

The imbalance issue in MMDG is more intricate than in standard multimodal learning due to its dual objectives: accurate classification and domain invariant generalization. Traditional imbalance mitigation methods focus mainly on improving classification within the source domain. However, this neglects a key insight: a modality that excels in source domain classification may still fail to learn domain invariant features effectively. As a result, balancing solely based on classification performance often has limited impact on generalization to target domains. This limitation is empirically demonstrated in Table \ref{tab:unimodal}, where traditional methods show improved source domain performance but limited generalization to unseen domains.

\begin{figure*}
    \centering
    \includegraphics[width=0.95\linewidth]{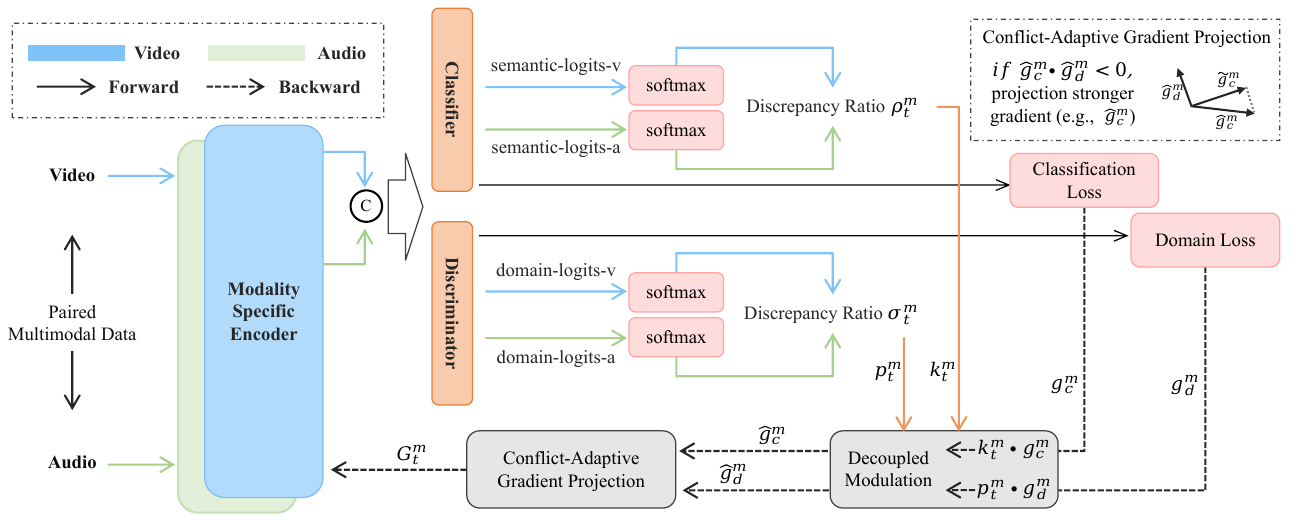}
    \caption{The overall training strategy of our proposed GMP.}
    \label{fig:pipeline}
\end{figure*}

To address these challenges, we propose a unified optimization strategy for MMDG, termed Gradient Modulation Projection (GMP) (Fig. \ref{fig:pipeline}), designed to fully exploit the domain generalization capabilities of each modality. GMP integrates two key components. The first component, Inter-Modality Gradient Decoupled Modulation (IGDM), differs from conventional methods by separating gradients for classification and domain invariance tasks. This decoupling allows the model to independently assess and optimize each modality’s contribution to both objectives. IGDM employs a dual modulation approach guided by two confidence metrics: semantic confidence for classification and domain confidence for generalization. These metrics enable dynamic, task specific gradient modulation based on each modality’s strengths. The second component, Conflict-Adaptive Gradient Projection (CAGP), addresses gradient conflicts within modality specific encoders. Through theoretical analysis, we show that classification and domain invariance gradients often point in opposing directions, impeding effective optimization. CAGP mitigates this by continuously monitoring task intensities during training and dynamically adjusting gradient projections. When conflicts arise, the relatively weaker task gradient is preserved, while the stronger one is projected orthogonally to reduce interference.

As shown in Table \ref{tab:unimodal}, existing balancing strategies such as OGM-GE \cite{Wang_2020_CVPR} and Grad-Blending \cite{peng2022balanced} achieve notable improvements on the source domain compared to the baseline model, but yield only marginal gains on the target domain (e.g., +0.65\% and +0.43\% in the video-audio task, respectively). In contrast, our proposed GMP achieves a similar improvement on the source domain while delivering a substantially higher gain on the target domain (+2.30\%). These results demonstrate that GMP maintains strong source domain accuracy without sacrificing generalization, promoting more balanced multimodal learning.

Our contributions are summarized as follows:
\begin{itemize}
\item We analyze why traditional multimodal learning balancing approaches fail to resolve optimization imbalance in MMDG. Empirically, we show that they often prioritize source domain classification, while neglecting generalization to unseen domains.

\item We introduce GMP, a unified strategy that adjusts classification and domain invariance gradients through IGDM and resolves inter-task conflicts using CAGP.

\item To the best of our knowledge, this is the first work to examine and address MMDG from an optimization perspective, offering a new path for MMDG task.

\item Extensive experiments demonstrate that our unified GMP strategy not only achieves state of the art performance but also integrates smoothly with diverse MMDG methods (Fig. \ref{fig:teaser}), showcasing strong versatility.
\end{itemize}

\begin{table}[t]
\centering
\setlength{\tabcolsep}{0pt}
\begin{tabular*}{\linewidth}{@{\extracolsep{\fill}}ccccc}
\midrule
Domain & Method & Video & Audio & Video-Audio \\
\midrule
\multirow{6}{*}{Source}
& Uni-video & 74.55 & - & - \\
& Uni-audio & - & 52.35 & - \\
\cmidrule(lr){2-5}
& Base & 70.28 & 48.63 & 76.31 \\
& OGM-GE & 72.66 & 50.33 & 77.69 \\
& Grad-Blending & 72.31 & 50.62 & 78.03 \\
& GMP(Ours) & 72.69 & 50.52 & 78.05 \\
\midrule
\multirow{6}{*}{Target}
& Uni-video & 54.98 & - & - \\
& Uni-audio & - & 38.86 & - \\
\cmidrule(lr){2-5}
& Base & 48.86 & 34.15 & 55.06 \\
& OGM-GE & 49.75 & 35.33 & 55.71 \\
& Grad-Blending & 50.03 & 35.26 & 55.49 \\
& GMP(Ours) & \textbf{52.33} & \textbf{35.88} & \textbf{57.36} \\
\midrule
\end{tabular*}
\caption{Performance comparison on source and target sets using the EPIC-Kitchens dataset \cite{kay2017kinetics}. Source results are averaged over validation sets (of the seen domains), and target results are averaged over test sets from the corresponding unseen domains. Uni-video and Uni-audio denote independently trained single-modality models.}
\label{tab:unimodal}
\end{table}

\section{Related Work}

\subsection{Multimodal Domain Generalization}
Unimodal domain generalization (DG) has been extensively studied, providing a solid foundation for understanding model generalization across domains. In contrast, MMDG remains in its early stages, with recent efforts focusing on multimodal alignment and representation learning \cite{li2025towards}. RNA-Net \cite{planamente2022domain} introduces a relative canonical alignment loss to balance audio and video features, tackling the alignment challenges of heterogeneous modalities. SimMMDG \cite{dong2023simmmdg} decouples modalities to better capture semantic structures and enhance domain invariant representations. MOOSA \cite{dong2024moosa} leverages an auxiliary pretext task to promote cross modal relationships, thereby improving feature representations. Similarly, CMRF \cite{fan2024cross} flattens cross modal representations to strengthen feature alignment and consistency. While these methods contribute to MMDG, they primarily emphasize domain invariant learning and overlook optimization dynamics between modalities, an underexplored yet crucial issue, as intermodal imbalance during training can hinder generalization.

\subsection{Imbalanced Multimodal Learning}
Multimodal models often favor modalities that are easier to learn, limiting the contribution of more complex ones \cite{ma2024tima,li2025taco,li2025miv,li2025context,li2025rrgmambaformer}. Despite access to diverse multimodal data, performance improvements remain constrained by inherent modality disparities, such as varying convergence speeds. To address this, several approaches seek to balance learning across modalities. Grad-Blending \cite{Wang_2020_CVPR} exploits overfitting tendencies to optimize modality combinations. OGM-GE \cite{peng2022balanced} builds on this by adaptively controlling gradients to maintain intermodal balance. AGM \cite{li2023boosting} uses Shapley values to adjust gradients based on single modal responses, encouraging equitable learning. CGGM \cite{guo2024classifier} regulates gradient magnitude and direction using classifier guided signals to address imbalance. DRB \cite{wei2024diagnosing} estimates each modality’s learning state through the separability of its single modal representation, then softly reinitializes encoders to prevent overfitting to less informative inputs. BALGRAD \cite{kwon2025see} further reduces intermodal differences through alignment to encourage balanced learning. Although these methods enhance performance on seen domains, they often underperform in MMDG settings due to a key limitation: they fail to account for each modality’s domain generalization capability, reducing robustness on unseen domains.

\section{Preliminaries}
\subsection{Training Setting}
The training dataset \(\mathcal{T} = \{x_i, y_i, d_i\}_{i=1}^N\) consists of samples from \(C\) source domains, where \(d_i \in \{1, \ldots, C\}\) denotes the domain label and \(y_i \in \{1, \ldots, Y\}\) the class label for each sample \(i\). Each input \(x_i = (x^v_i, x^a_i)\) includes video (\(m = v\)) and audio (\(m = a\)) modalities. The objective is to learn class-informative, domain invariant features that generalize to unseen domains by balancing two goals: (1) classification, which promotes class-informative features, and (2) domain invariance, which mitigates domain specific biases. To this end, we adopt a combined loss function used in unimodal domain generalization \cite{li2018domain} and adapt it to the multimodal case, where the classification loss \(L_c\) and domain adversarial loss \(L_d\) are computed via a classifier and a domain discriminator, respectively:
\begin{equation}
L = L_c + \lambda L_d,
\end{equation}
where \(\lambda\) controls the trade-off between the two objectives. At training step \(t\), let \(\theta_t^m\) denote the parameters for modality \(m \in \{v, a\}\). The classification and domain gradients with respect to \(\theta_t^m\) are $\nabla_{\theta_t^m} L_c$ and $\nabla_{\theta_t^m} L_d$, respectively. Incorporating trade-off \(\lambda\), we define the gradients for modality $m$ at step $t$:
\begin{equation}
g_{c}^m = \nabla_{\theta_t^m} L_c, \quad g_{d}^m = (- \lambda) \cdot \nabla_{\theta_t^m} L_d.
\end{equation}
where the negative sign in $g_{d}^m$ implements gradient reversal. The parameter update:
\begin{equation}
\begin{aligned}
\theta^{m}_{t+1} 
& = \theta^m_{t} - \eta \left( \nabla_{\theta^m} L_c -\lambda \nabla_{\theta^m} L_d \right) \\
& = \theta^m_{t} - \eta \left( g_{c}^m + g_{d}^m \right) \\
& = \theta^m_{t} - \eta  G_t^m ,
\end{aligned}
\end{equation}
where $G_t^m$ denotes the total gradient applied to modality $m$  at step $t$. We now analyze the unbalanced forms in MMDG: inter-modality imbalance and inter-task imbalance.

\subsection{Inter-Modality Imbalance} 
In multimodal learning, persistent discrepancies in gradient magnitudes across modalities can result in one modality dominating the optimization process, thereby diminishing contributions of others \cite{peng2022balanced}. Specifically, if $\left| g_{c}^v \right| \gg \left| g_{c}^a \right|$, the video modality exerts a disproportionately strong influence. This dominance is quantified by the ratio $r_{v,a}(t) = || g_{c}^v || / || g_{c}^a || \gg 1$ indicating that the video gradients dominate the updates at step $t$. Furthermore, this imbalance accumulates significantly over $T$ steps as $R_{v,a}(T) = (\sum_{t=1}^T \| g_{c}^v \|) / (\sum_{t=1}^T \| g_{c}^a \|)$, which leads to chronic under optimization of weaker modalities like audio.

However, MMDG introduces a critical nuance in which conventional gradient modulation strategies that balance modalities based solely on classification performance (reflected in $\left| g_{c}^m \right|$) prove fundamentally inadequate. This inadequacy arises for several reasons. There is a generalization critical asymmetry where a modality with strong source domain discriminative power may exhibit weak domain invariance, whereas a seemingly weaker modality could excel at capturing domain agnostic features essential for unseen domains. Therefore, simply balancing via $R_{v,a}(T)$,  which ignores invariance quality, risks suppressing modalities that are vital for cross domain generalization. As shown in Table \ref{tab:unimodal}, traditional balancing approaches often fail on MMDG benchmarks because they favor source domain accuracy over robust generalization.

\subsection{Inter-Task Conflicts}
Another challenge arises when gradients from different tasks, such as classification and domain adversarial training, conflict with each other and exhibit negative cosine similarity \cite{yu2020gradient}. In such cases, a dominant gradient may overshadow others, degrading the performance of less represented tasks. In DG tasks, this issue often arises when the classification loss gradient $g_{c}^m$ and the domain adversarial loss gradient $g_{d}^m$ point in opposing directions.

Let ${g}^m_{c} = \{g_{c}^{v},g_{c}^{a}\}$ and ${g}^{m}_{d} = \{g_{d}^{v},g_{d}^{a}\}$ represent the gradients of the classification loss $L_c$ and the domain adversarial loss $L_d$, respectively, where $\theta = [\theta^v, \theta^a]$ are the modality specific parameters for video and audio. Assuming updates are performed using gradient descent, the parameter updates become:

$$
\theta^v_{t+1} = \theta^v_t - \eta \left(g_{c}^{v} + g_{d}^{v}\right), \quad \theta^a_{t+1} = \theta^a_t - \eta \left(g_{c}^{a} + g_{d}^{a}\right).
$$

For the combined loss $L = L_c + \lambda L_d$, the first order approximation of the change in total loss after a single update step is:

\begin{equation}
\begin{aligned}
\Delta L &= L(\theta_{t+1}) - L(\theta_t) \\
&= -\eta \left( || {g}^m_{c} ||^2 + || {g}^{m}_{d} ||^2 + 2 {{g}^m_c}^\top {g}^{m}_{d} \right)+ \mathcal{O}(\eta^2).
\end{aligned}
\end{equation}
If ${{g}^m_c}^\top {g}^{m}_{d} < 0$, the classification and domain adversarial gradients are in conflict, which misalignment can impede convergence and deteriorate multi task optimization.

In MMDG, such conflicts are often modality specific. For instance, video related gradients may show stronger conflict between classification and domain objectives, while audio gradients remain more aligned. Consequently, applying a uniform conflict resolution strategy across modalities fails to capture these modality specific dynamics. As demonstrated in Table~\ref{tab:ablation}, traditional conflict mitigation strategies underperform, ultimately limiting generalization to unseen domains.

\section{Method}
\subsection{Inter-Modality Gradient Decoupled Modulation}
To address inter-modality imbalance, we propose an Inter-Modality Gradient Decoupled Modulation (IGDM) strategy for MMDG. Unlike traditional unified gradient balancing methods that mainly consider overall modality classification ability, IGDM introduces a decoupled modulation strategy that adjusts each modality’s contribution independently for classification and domain discrimination tasks. This fine grained control directly tackles the shortcomings of classification gradient only balancing strategies.

We consider video and audio modalities, each processed by a modality specific encoder $\varphi^m(\theta^m, \cdot)$ that extracts features from the input. These features are concatenated and fed into a classifier $f_c$ and a domain discriminator $f_d$:
\begin{equation}
\begin{aligned}
f_c(x_i) & = \sum_{m=1}^M W^m \varphi^m(\theta^m, x^m_i) + b_y, \\
f_d(x_i) & = \sum_{m=1}^M D^m \varphi^m(\theta^m, x^m_i) + b_d,
\end{aligned}
\end{equation}
where \(M\) is the number of modalities, $W^m$ and $D^m$ are the respective weights for classification and domain discrimination, and $b_y$, $b_d$ are the bias terms.

As established in prior work \cite{peng2022balanced}, modalities with higher prediction confidence tend to contribute less to the joint gradient through their respective weights. This results in dominant modalities suppressing gradient updates for weaker ones, leading to biased optimization. To quantify individual modality contributions and counteract this effect, we introduce two distinct confidence metrics. Semantic Confidence measures classification certainty; higher values indicate stronger discriminative power. Domain Confidence measures domain confusion; lower values indicate stronger domain invariant features.

For each modality $m$ and sample $i$, we compute semantic confidence $q^{m}_i$ using its contribution from the classifier:
\begin{equation}
q^{m}_i = \sum_{k=1}^Y 1_{k=y_{i}} \cdot \text{softmax} \left( W^{m}_{t} \cdot \varphi^{m}_{t}(\theta^{m}, x_{i}^{m}) \right)_k,
\end{equation}
where $Y$ is the number of classes. Domain confidence $c^{m}_i$ is similarly derived from the domain discriminator's output:
\begin{equation}
c^{m}_i = \sum_{k=1}^C 1_{k=d_{i}} \cdot \text{softmax} \left( D^{m}_{t} \cdot \varphi^{m}_{t}(\theta^{m}, x_{i}^{m}) \right)_k,
\end{equation}
where $C$ is the number of domains.

To compare modalities within a mini-batch $B_t$, we define two discrepancy ratios:
\begin{equation}
\rho^{m}_{t} = \frac{\sum_{i \in B_{t}} q^{m}_i}{\sum_{i \in B_{t}} q^{\bar{m}}_i}, \quad
\sigma^{m}_{t} = \frac{\sum_{i \in B_{t}} c^{\bar{m}}_i}{\sum_{i \in B_{t}} c^{m}_i},
\end{equation}
where $\bar{m}$ denotes the other modality. These ratios are defined in complementary directions: high semantic confidence benefits classification (higher numerator), while low domain confidence benefits domain generalization (lower denominator). Hence, $\rho^{m}_{t} > 1$ indicates that modality $m$ is stronger in classification, while $\sigma^{m}_{t} > 1$ implies that modality $m$ is stronger in domain invariance.

To modulate gradient flow accordingly, we introduce two independent, decoupled modulation coefficients, \(k^{m}_{t}\) and \(p^{m}_{t}\), for classification and domain discrimination gradients, respectively:
\begin{equation}
k^{m}_{t} = \begin{cases}
1 - \tanh (\alpha_k \cdot \rho^{m}_{t}) & \text{if } \rho^{m}_{t} > 1 \\
1 & \text{otherwise},
\end{cases}
\end{equation}
\begin{equation}
p^{m}_{t} = \begin{cases}
1 - \tanh (\alpha_p \cdot \sigma^{m}_{t}) & \text{if } \sigma^{m}_{t} > 1 \\
1 & \text{otherwise},
\end{cases}
\end{equation}
where \(\alpha_k, \alpha_p\) are hyperparameters controlling suppression strength. The \(\tanh\) function ensures coefficients are bounded in \([0, 1]\), maintaining training stability. 

This decoupled strategy enables the model to scale down the classification gradient $g_{c}^m$ using $k^{m}_{t}$ and the domain gradient $g_{d}^m$ using $p^{m}_{t}$, independently. For instance, when modality $m$ is already strong in classification ($\rho^{m}_{t} > 1$), $k^{m}_{t}$ reduces its classification gradient, allowing weaker modalities to contribute more. A similar principle holds for domain gradients when $\sigma^{m}_{t} > 1$.

The final modulated gradients are:
\begin{equation}
\hat{g}_{c}^m = k^{m}_{t} \cdot g_{c}^m , \quad \hat{g}_{d}^m = p^{m}_{t} \cdot g_{d}^m .
\end{equation}

\subsection{Conflict-Adaptive Gradient Projection}
To mitigate inter-task conflicts during optimization, we introduce the Conflict-Adaptive Gradient Projection (CAGP) strategy, which explicitly addresses gradient interference between classification and domain invariant learning objectives. In MMDG, gradient conflicts between these objectives, indicated by \(\hat{g}_{c}^m \cdot \hat{g}_{d}^m < 0\), are often highly asymmetric across modalities. For example, such conflicts tend to be severe in video (\(m=v\)) but minimal in audio (\(m=a\)). Additionally, their relative task strengths vary depending on the modality and over time, making uniform conflict resolution strategies suboptimal.

CAGP tackles these limitations through three strategys. First, it is conflict-aware, projecting only when gradient conflicts arise, which helps maintain task synergy. Second, it is modality specific, projecting independently within each modality to capture distinct dynamics. Third, it is task strength adaptive, favoring updates that support the weaker task to ensure balanced learning and avoid task dominance. Its purpose is to remove conflicting components from the update direction, preserving useful progress while avoiding destructive interference. If \(\hat{g}_{c}^m \cdot \hat{g}_{d}^m \geq 0\), the gradients are aligned or orthogonal, and no action is needed. To avoid weakening the less optimized task, we prioritize its full update direction. Thus, when a conflict is detected, we resolve it by projecting the stronger task’s gradient orthogonal to the weaker one, thereby preserving progress for the weaker objective.

To determine which task is stronger, we use the relative task strength ratio $\Gamma_t^m = {\rho_t^m}/{\sigma_t^m}$, where $\rho_t^m$ reflects the classification strength of modality $m$ (via high semantic confidence), and $\sigma_t^m$ reflects its domain invariance strength (via low domain confidence). 

If $\Gamma_t^m > 1$, classification is the relatively stronger task. We project its gradient away from the direction of the domain invariant gradient:
\begin{equation}
\tilde{g}^{m}_{c} = \hat{g}_{c}^m - \frac{ \hat{g}_{c}^m \cdot \hat{g}_{d}^m }{||\hat{g}_{d}^m||^2} \hat{g}_{d}^m.
\end{equation}
This projects \(g^m_{c}\) orthogonally to \(g^m_{d}\) to remove the conflicting component with the domain task. The updated gradient becomes:
  \begin{equation}
  G^{m}_{t} = \tilde{g}^{m}_{c} + \hat{g}_{d}^m.
  \end{equation}
Conversely, if \(\Gamma_t^m < 1\), classification is relatively weaker, and we project the domain invariant learning gradient onto a direction orthogonal to the classification task:
\begin{equation}
\tilde{g}^{m}_{d} = \hat{g}_{d}^m - \frac{ \hat{g}_{d}^m  \cdot \hat{g}_{c}^m}{||\hat{g}_{c}^m||^2} \hat{g}_{c}^m,
\end{equation}
 and update with:
\begin{equation}
G^{m}_{t} = \hat{g}_{c}^m + \tilde{g}^{m}_{d}.
\end{equation}

This projection strategy ensures that the relatively weaker task’s full gradient is preserved, while only the conflicting component of the stronger task is removed.

\section{Experiments}
\noindent\textbf{Datasets and Implementation Details.}
We evaluate on two popular multimodal benchmarks: EPIC-Kitchens \cite{damen2018scaling} and HAC \cite{dong2023simmmdg}, both offering synchronized video and audio. Following \cite{dong2023simmmdg}, we use a domain generalization setup: training on two domains and testing on a third, reporting the average performance across all unseen target domains. Our setup is based on \cite{dong2023simmmdg} and implemented with MMAction2 \cite{2020mmaction2}. We use SlowFast for video and ResNet-18 for audio, pretrained on Kinetics-400 \cite{kay2017kinetics} and VGGSound \cite{chen2020vggsound}, respectively. Models are trained using Adam (learning rate 1e-4) for 20 epochs on a RTX 4090 GPU. The best model is selected based on validation performance, with final results averaged over five runs. Both the domain discriminator and classifier are MLPs with 256 dimensional hidden layers. Hyperparameters are tuned in [0,1] using validation data.

\begin{table}[t]
\centering
\begin{tabular}{c|cc}
\toprule
Method & EPIC-Kitchens &  HAC \\
\midrule
Base & 55.06 & 61.86 \\
Grad-Blending & 55.49 & 62.66 \\
OGM-GE & 55.71 & 62.83 \\
AGM & 55.39 & 62.16 \\
CGGM & 55.30 & 62.80 \\
DRB & 54.88 & 61.92 \\
BALGRAD & 55.26 & 62.36 \\
\midrule
\rowcolor{gray!20}\textbf{GMP(Ours)} & \textbf{57.36} & \textbf{64.91} \\  
\toprule
\end{tabular}
\caption{Comparison of our proposed GMP strategy with existing gradient strategies. All methods use concatenation fusion as the base.}
\label{tab:gradient}
\end{table}

\noindent\textbf{Comparison with Other Gradient Strategies.}
We compare our method with several state of the art gradient modulation strategies tailored for multimodal learning, including Grad-Blending \cite{Wang_2020_CVPR}, OGM-GE \cite{peng2022balanced}, AGM \cite{li2023boosting}, CGGM \cite{guo2024classifier}, DRB \cite{wei2024diagnosing}, and BALGRAD \cite{kwon2025see}. To ensure fair comparison, we reimplement all methods using the same architecture and training protocol described above. As shown in Table~\ref{tab:gradient}, although existing strategies alleviate gradient imbalance and outperform the naive Base method, they are less effective for MMDG. In contrast, our proposed GMP strategy consistently outperforms all baselines on both EPIC-Kitchens and HAC datasets. These results underscore the need for MMDG specific gradient optimization methods.

\begin{table}[t]
\centering
\begin{tabular}{c|cc}
\toprule
Method & EPIC-Kitchens &  HAC \\
 \midrule
RNA-Net & 60.31 & 62.88 \\
MOOSA & 60.29 & 67.39 \\
SimMMDG & 60.81 & 67.82 \\
CMRF & 60.60 & 67.66 \\
\midrule
\rowcolor{gray!20} RNA-Net\dag & 60.92 & 65.21 \\
\rowcolor{gray!20} MOOSA\dag & 61.79 & 68.75 \\
\rowcolor{gray!20} SimMMDG\dag & 62.03 & \textbf{69.11} \\
\rowcolor{gray!20} CMRF\dag & \textbf{62.30} & 68.93 \\
\midrule
\end{tabular}
\caption{Integration with existing MMDG methods. Methods marked with \dag~indicate integration of GMP.}
\label{tab:MMDG}
\end{table}

\noindent\textbf{Integration with Existing MMDG Methods.}
To evaluate the compatibility of our optimization strategy with existing MMDG frameworks, we integrate GMP into several representative methods: RNA-NET \cite{planamente2024relative}, MOOSA \cite{dong2024moosa}, SimMMDG \cite{dong2023simmmdg}, and CMRF \cite{fan2024cross}. These baselines primarily target architectural or representational improvements to handle modality and domain gaps. As shown in Table~\ref{tab:MMDG}, incorporating GMP consistently yields further performance gains across all cases. This demonstrates that our approach functions as a plug-and-play optimization module, enhancing generalization capabilities in unseen domains.

\begin{table}[t]
\centering
\setlength{\tabcolsep}{4mm}{
\begin{tabular}{c|cc}
\toprule
Method & EPIC-Kitchens &  HAC \\
 \midrule
Base & 55.06 & 61.86 \\
IGDM-only & 55.98 & 63.05 \\
CAGP-only & 55.34 & 63.41 \\
\rowcolor{gray!20} Full & \textbf{57.36} & \textbf{64.91} \\
 \midrule
Unified Modulation & 54.97 & 62.50 \\
w/o $ k_t^m $ & 55.19 & 62.33 \\
w/o $ p_t^m $ & 55.70 & 62.29 \\
\rowcolor{gray!20} Full IGDM & \textbf{55.98} & \textbf{63.05} \\
 \midrule
Fixed Proj-Class & 54.63 & 62.38 \\
Fixed Proj-Domain & 53.88 & 62.03 \\
PCGrad & 54.33 & 62.64 \\
Reverse CAGP & 54.00 & 62.57 \\
\rowcolor{gray!20} Full CAGP & \textbf{55.34} & \textbf{63.41} \\
 \midrule
\end{tabular}}
\caption{Ablation study results demonstrating the effectiveness of individual components and strategies.}
\label{tab:ablation}
\end{table}

\noindent\textbf{Uni-modal Performance Comparison.}
As shown in Table~\ref{tab:gradient}, our method substantially improves single-modality generalization to unseen domains. The video branch achieves 52.33\% accuracy (a +3.47\% gain over the baseline), while the audio branch reaches 35.88\% (+1.73\%). These results indicate that GMP successfully unlocks the generalization capacity of each modality. Notably, traditional joint training often suppresses unimodal capabilities; for instance, the baseline video branch performs 6.12\% worse than its independently trained counterpart. GMP reduces this gap to just 2.65\%.

\begin{figure}
    \centering
    \includegraphics[width=0.95\linewidth]{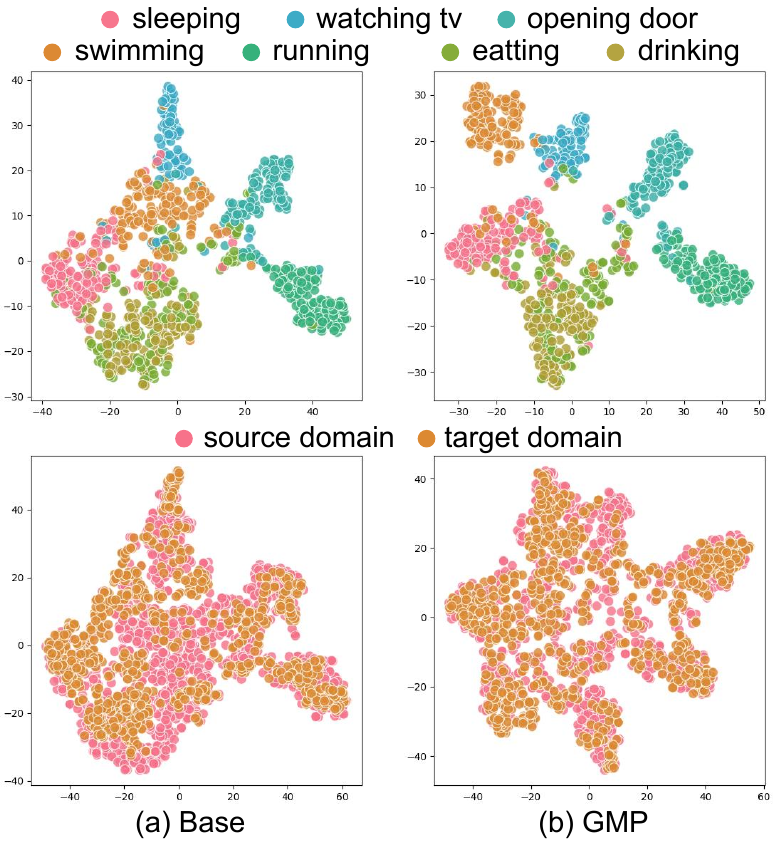}
    \caption{t-SNE plots on the HAC dataset (target domain A) compare the Base and GMP methods in classification and domain invariance.}
    \label{fig:tsne}
\end{figure}

\noindent\textbf{Ablation Study.}
We perform extensive ablation studies to assess the contribution of each proposed component. Quantitative results are presented in Table~\ref{tab:ablation}.

\noindent\textit{Effectiveness of IGDM and CAGP:}
Adding IGDM or CAGP individually to the baseline yields noticeable improvements on both EPIC and HAC datasets. When combined, the full model achieves the highest performance, confirming that these components offer complementary benefits.

\noindent\textit{Decoupled vs. Unified Modulation:}
Replacing IGDM’s decoupled strategy with a unified gradient scaling scheme leads to a performance drop. Further analysis shows that disabling either semantic confidence modulation ($k_t^m$) or domain confidence modulation ($p_t^m$) results in significant accuracy loss. These findings emphasize the importance of jointly maintaining both forms of confidence-based modulation for effective modality balancing.

\noindent\textit{Adaptive Projection vs. Alternatives:}
Replacing CAGP with fixed projection strategies that target either classification or domain gradients results in inferior performance. Although PCGrad \cite{yu2020gradient} partially alleviates conflicts, it still underperforms compared to CAGP. Notably, Reverse CAGP, which protects the stronger task rather than the weaker one, also performs worse. These comparisons confirm the design choice of protecting the weaker task through task strength adaptive projection, as this approach more effectively resolves gradient conflicts and improves performance.

\begin{table}
\centering
\setlength{\tabcolsep}{4mm}{
\begin{tabular}{c|cc}
\toprule
Method & EPIC-Kitchens &  HAC \\
 \midrule
Activation-based & 52.23 & 57.49 \\
Concatenation & 55.06 & 61.86 \\
Summation & 52.94 & 58.51 \\
FiLM & 53.16 & 59.35 \\ 
 \midrule
Activation-based\dag & 55.69 & 61.33 \\
\rowcolor{gray!20} Concatenation\dag & \textbf{57.36} & \textbf{64.91} \\
Summation\dag & 56.37 & 63.21 \\
FiLM\dag & 56.19 & 63.04 \\
 \midrule
\end{tabular}}
\caption{Combined with GMP, conventional fusion methods consistently gain considerable improvement. \dag~ indicates GMP strategy is applied.}
\label{tab:conventional}
\end{table}

\begin{figure}
    \centering
    \includegraphics[width=0.95\linewidth]{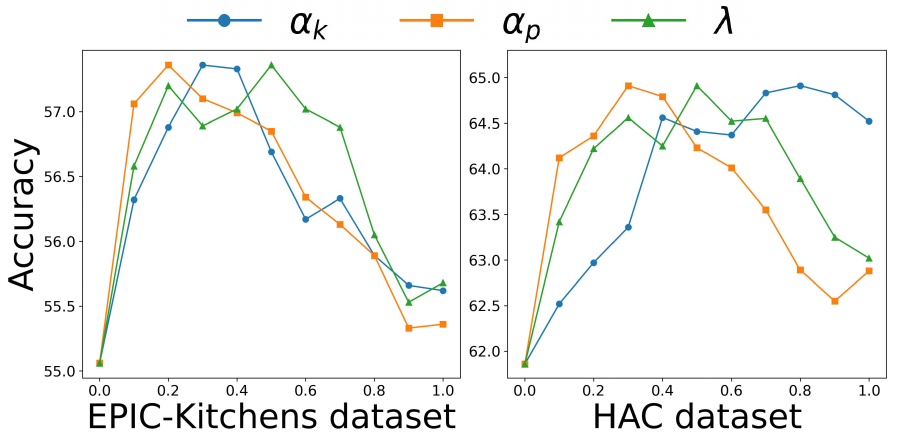}
    \caption{Hyperparameter sensitivity analysis of GMP. }
    \label{fig:hyper}
\end{figure}

\noindent\textbf{Visualization.}
To further validate our approach, we present a t-SNE visualization on the HAC dataset (target domain A) in Fig. \ref{fig:tsne}. As observed, the baseline method suffers from overlapping class clusters and a distinct distribution gap between domains. Conversely, our GMP method produces well-separated decision boundaries with distinct class structures. Furthermore, GMP demonstrates superior domain invariance, evidenced by the significant alignment between source and target features. This qualitative analysis confirms that our dual-objective strategy effectively learns representations that are both class-discriminative and domain-invariant, ensuring robust generalization.

\noindent\textbf{Integration with Conventional Fusion Methods.}
To further demonstrate the versatility of our approach, we integrate GMP with various conventional fusion strategies, including summation, concatenation, activation-based fusion, and the adaptive FiLM mechanism \cite{perez2018film}. As shown in Table~\ref{tab:conventional}, our method consistently improves performance across all fusion types. These results demonstrate that GMP is agnostic to fusion paradigms and complements a wide range of multimodal fusion schemes, enhancing generalization without reliance on any specific fusion design.

\begin{figure}
    \centering
    \includegraphics[width=1.0\linewidth]{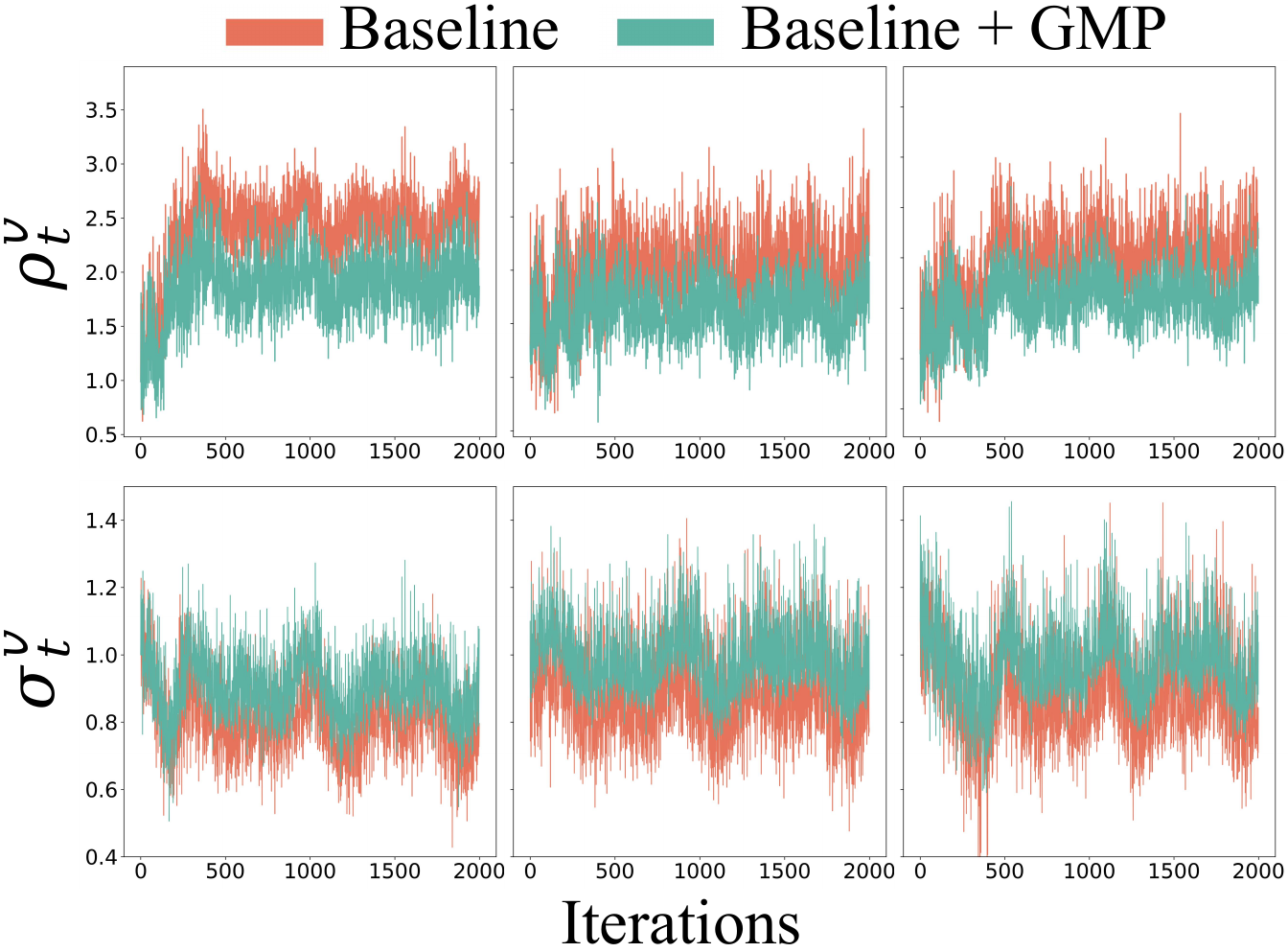}
    \caption{Discrepancy ratio trajectories ($\rho_t^m$, $\sigma_t^m$) on EPIC-Kitchens. GMP mitigates cross-modal imbalance by driving ratios toward 1, achieving balanced semantic and domain optimization.}
    \label{fig:ratio_curve}
\end{figure}

\noindent\textbf{Hyperparameter Sensitivity.}
We conduct controlled experiments to evaluate the sensitivity of GMP to its hyperparameters $\alpha_k$, $\alpha_p$, and $\lambda$. The parameters $\alpha_k$ and $\alpha_p$ control the suppression strength of dominant modalities in the classification and domain branches, respectively, while $\lambda$ balances classification accuracy with domain invariance. Setting $\alpha_k$ and $\alpha_p$ to 0.0 disables suppression, whereas higher values apply stronger attenuation. As shown in Fig. \ref{fig:hyper}, most settings consistently improve performance over the baseline, indicating that calibrated suppression reduces modality dominance and improves generalization without diminishing the influence of strong modalities.

\noindent\textbf{Discrepancy Ratio Analysis.}
To quantify the balance between modalities during training, we define two discrepancy ratios: the semantic discrepancy ratio ($\rho_t^m$) and the domain discrepancy ratio ($\sigma_t^m$), as described in Eq. (8). These ratios measure the relative semantic discriminativeness and domain invariance of modality $m$ with respect to its counterpart $\bar{m}$ at training step $t$. Ideally, both values converge to 1, reflecting balanced contributions. Significant deviations indicate optimization asymmetry, where one modality dominates, leading to underutilization of the other. As illustrated in Fig.~\ref{fig:ratio_curve}, training without our method results in persistent imbalance. In contrast, our GMP reduces this discrepancy by guiding both $\rho_t^m$ and $\sigma_t^m$ closer to 1, thereby promoting balanced inter-modality learning.

\section{Conclusion}
We propose Gradient Modulation Projection (GMP), a unified optimization strategy for Multimodal Domain Generalization (MMDG). GMP addresses gradient imbalances and task conflicts in multimodal learning, enabling models to better generalize to unseen domains. Experiments on multiple datasets demonstrate state-of-the-art generalization performance. GMP is architecture-agnostic, consistently enhances existing MMDG methods, and offers the first optimization-centric solution for multimodal domain shifts.

\section{Acknowledgments}
This work was supported by the National Key R\&D Program of China (2024YFB3311600) and the Natural Science Foundation of Henan (252300423936).

\bibliography{aaai2026}

\begin{thebibliography}{33}
\providecommand{\natexlab}[1]{#1}

\bibitem[{Chen et~al.(2020)Chen, Xie, Vedaldi, and Zisserman}]{chen2020vggsound}
Chen, H.; Xie, W.; Vedaldi, A.; and Zisserman, A. 2020.
\newblock Vggsound: A large-scale audio-visual dataset.
\newblock In \emph{ICASSP 2020-2020 IEEE International Conference on Acoustics, Speech and Signal Processing (ICASSP)}, 721--725. IEEE.

\bibitem[{Contributors(2020)}]{2020mmaction2}
Contributors, M. 2020.
\newblock OpenMMLab's Next Generation Video Understanding Toolbox and Benchmark.
\newblock \url{https://github.com/open-mmlab/mmaction2}.

\bibitem[{Damen et~al.(2018)Damen, Doughty, Farinella, Fidler, Furnari, Kazakos, Moltisanti, Munro, Perrett, Price et~al.}]{damen2018scaling}
Damen, D.; Doughty, H.; Farinella, G.~M.; Fidler, S.; Furnari, A.; Kazakos, E.; Moltisanti, D.; Munro, J.; Perrett, T.; Price, W.; et~al. 2018.
\newblock Scaling egocentric vision: The epic-kitchens dataset.
\newblock In \emph{Proceedings of the European conference on computer vision (ECCV)}, 720--736.

\bibitem[{Dong, Chatzi, and Fink(2024)}]{dong2024moosa}
Dong, H.; Chatzi, E.; and Fink, O. 2024.
\newblock Towards Multimodal Open-Set Domain Generalization and Adaptation through Self-supervision.
\newblock In \emph{European Conference on Computer Vision}.

\bibitem[{Dong et~al.(2025{\natexlab{a}})Dong, Liu, Zhou, Chatzi, Kannala, Stachniss, and Fink}]{dong2025mmdasurvey}
Dong, H.; Liu, M.; Zhou, K.; Chatzi, E.; Kannala, J.; Stachniss, C.; and Fink, O. 2025{\natexlab{a}}.
\newblock Advances in Multimodal Adaptation and Generalization: From Traditional Approaches to Foundation Models.
\newblock \emph{arXiv preprint arXiv:2501.18592}.

\bibitem[{Dong et~al.(2023)Dong, Nejjar, Sun, Chatzi, and Fink}]{dong2023simmmdg}
Dong, H.; Nejjar, I.; Sun, H.; Chatzi, E.; and Fink, O. 2023.
\newblock SimMMDG: A simple and effective framework for multi-modal domain generalization.
\newblock \emph{Advances in Neural Information Processing Systems}, 36: 78674--78695.

\bibitem[{Dong et~al.(2025{\natexlab{b}})Dong, Sheng, Liang, He, Chatzi, and Fink}]{dong2025adapting}
Dong, H.; Sheng, L.; Liang, J.; He, R.; Chatzi, E.; and Fink, O. 2025{\natexlab{b}}.
\newblock Adapting Vision-Language Models Without Labels: A Comprehensive Survey.
\newblock \emph{arXiv preprint arXiv:2508.05547}.

\bibitem[{Fan et~al.(2024)Fan, Xu, Wang, and Guo}]{fan2024cross}
Fan, Y.; Xu, W.; Wang, H.; and Guo, S. 2024.
\newblock Cross-modal representation flattening for multi-modal domain generalization.
\newblock \emph{Advances in Neural Information Processing Systems}, 37: 66773--66795.

\bibitem[{Galappaththige et~al.(2024)Galappaththige, Baliah, Gunawardhana, and Khan}]{galappaththige2024towards}
Galappaththige, C.~J.; Baliah, S.; Gunawardhana, M.; and Khan, M.~H. 2024.
\newblock Towards Generalizing to Unseen Domains with Few Labels.
\newblock In \emph{Proceedings of the IEEE/CVF Conference on Computer Vision and Pattern Recognition}, 23691--23700.

\bibitem[{Galappaththige, Kuruppu, and Khan(2024)}]{galappaththige2024generalizing}
Galappaththige, C.~J.; Kuruppu, G.; and Khan, M.~H. 2024.
\newblock Generalizing to unseen domains in diabetic retinopathy classification.
\newblock In \emph{Proceedings of the IEEE/CVF Winter Conference on Applications of Computer Vision}, 7685--7695.

\bibitem[{Guo et~al.(2024)Guo, Jin, Chen, and Zhao}]{guo2024classifier}
Guo, Z.; Jin, T.; Chen, J.; and Zhao, Z. 2024.
\newblock Classifier-guided gradient modulation for enhanced multimodal learning.
\newblock \emph{Advances in Neural Information Processing Systems}, 37: 133328--133344.

\bibitem[{Kay et~al.(2017)Kay, Carreira, Simonyan, Zhang, Hillier, Vijayanarasimhan, Viola, Green, Back, Natsev et~al.}]{kay2017kinetics}
Kay, W.; Carreira, J.; Simonyan, K.; Zhang, B.; Hillier, C.; Vijayanarasimhan, S.; Viola, F.; Green, T.; Back, T.; Natsev, P.; et~al. 2017.
\newblock The kinetics human action video dataset.
\newblock \emph{arXiv preprint arXiv:1705.06950}.

\bibitem[{Khan, Shaaban, and Khan(2024)}]{khan2024improving}
Khan, A.; Shaaban, M.~A.; and Khan, M.~H. 2024.
\newblock Improving pseudo-labelling and enhancing robustness for semi-supervised domain generalization.
\newblock \emph{arXiv preprint arXiv:2401.13965}.

\bibitem[{Kwon et~al.(2025)Kwon, Kim, Lee, Choi, and Kim}]{kwon2025see}
Kwon, J.; Kim, M.; Lee, E.; Choi, J.; and Kim, Y. 2025.
\newblock See-Saw Modality Balance: See Gradient, and Sew Impaired Vision-Language Balance to Mitigate Dominant Modality Bias.
\newblock In \emph{Proceedings of the 2025 Conference of the Nations of the Americas Chapter of the Association for Computational Linguistics: Human Language Technologies (Volume 1: Long Papers)}, 4364--4378.

\bibitem[{Li et~al.(2026)Li, Dong, Wan, Li, Xu, and Khan}]{li2026multimodaldomaingeneralizationlabels}
Li, H.; Dong, H.; Wan, H.; Li, S.; Xu, M.; and Khan, M.~H. 2026.
\newblock Towards Multimodal Domain Generalization with Few Labels.
\newblock arXiv:2602.22917.

\bibitem[{Li et~al.(2023)Li, Li, Hu, Lei, Li, and Zhou}]{li2023boosting}
Li, H.; Li, X.; Hu, P.; Lei, Y.; Li, C.; and Zhou, Y. 2023.
\newblock Boosting multi-modal model performance with adaptive gradient modulation.
\newblock In \emph{Proceedings of the IEEE/CVF International Conference on Computer Vision}, 22214--22224.

\bibitem[{Li et~al.(2025{\natexlab{a}})Li, Liu, Wang, Jiang, Jiu, Chen, Lu, Li, and Xu}]{li2025rrgmambaformer}
Li, H.; Liu, S.; Wang, H.; Jiang, X.; Jiu, M.; Chen, L.; Lu, Y.; Li, S.; and Xu, M. 2025{\natexlab{a}}.
\newblock RRGMambaFormer: A hybrid Transformer-Mamba architecture for radiology report generation.
\newblock \emph{Expert Systems with Applications}, 279: 127419.

\bibitem[{Li et~al.(2018)Li, Pan, Wang, and Kot}]{li2018domain}
Li, H.; Pan, S.~J.; Wang, S.; and Kot, A.~C. 2018.
\newblock Domain generalization with adversarial feature learning.
\newblock In \emph{Proceedings of the IEEE conference on computer vision and pattern recognition}, 5400--5409.

\bibitem[{Li et~al.(2025{\natexlab{b}})Li, Wan, Zhang, Jiu, Li, Xu, and Khan}]{li2025towards}
Li, H.; Wan, H.; Zhang, L.; Jiu, M.; Li, S.; Xu, M.; and Khan, M.~H. 2025{\natexlab{b}}.
\newblock Towards Robust Multimodal Domain Generalization via Modality-Domain Joint Adversarial Training.
\newblock In \emph{Proceedings of the 33rd ACM International Conference on Multimedia}, 180--188.

\bibitem[{Li et~al.(2025{\natexlab{c}})Li, Wang, Sun, He, and Feng}]{li2025context}
Li, H.; Wang, H.; Sun, X.; He, H.; and Feng, J. 2025{\natexlab{c}}.
\newblock Context-enhanced framework for medical image report generation using multimodal contexts.
\newblock \emph{Knowledge-Based Systems}, 310: 112913.

\bibitem[{Li et~al.(2025{\natexlab{d}})Li, Cao, He, Cheng, Fu, Xiao, Wang, and Tang}]{li2025miv}
Li, Y.; Cao, Y.; He, H.; Cheng, Q.; Fu, X.; Xiao, X.; Wang, T.; and Tang, R. 2025{\natexlab{d}}.
\newblock M{\texttwosuperior}{IV}: Towards Efficient and Fine-grained Multimodal In-Context Learning via Representation Engineering.
\newblock In \emph{Second Conference on Language Modeling}.

\bibitem[{Li et~al.(2025{\natexlab{e}})Li, Yang, Yun, Feng, Huang, and Tang}]{li2025taco}
Li, Y.; Yang, J.; Yun, T.; Feng, P.; Huang, J.; and Tang, R. 2025{\natexlab{e}}.
\newblock Taco: Enhancing multimodal in-context learning via task mapping-guided sequence configuration.
\newblock In \emph{Proceedings of the 2025 Conference on Empirical Methods in Natural Language Processing}, 736--763.

\bibitem[{Ma, Liu, and Cheng(2024)}]{ma2024tima}
Ma, F.; Liu, L.; and Cheng, H.~V. 2024.
\newblock TIMA: Text-Image Mutual Awareness for Balancing Zero-Shot Adversarial Robustness and Generalization Ability.
\newblock \emph{arXiv preprint arXiv:2405.17678}.

\bibitem[{Munir et~al.(2023)Munir, Khan, Sarfraz, and Ali}]{munir2023domain}
Munir, M.~A.; Khan, M.~H.; Sarfraz, M.~S.; and Ali, M. 2023.
\newblock Domain adaptive object detection via balancing between self-training and adversarial learning.
\newblock \emph{IEEE Transactions on Pattern Analysis and Machine Intelligence}, 45(12): 14353--14365.

\bibitem[{Peng et~al.(2022)Peng, Wei, Deng, Wang, and Hu}]{peng2022balanced}
Peng, X.; Wei, Y.; Deng, A.; Wang, D.; and Hu, D. 2022.
\newblock Balanced multimodal learning via on-the-fly gradient modulation.
\newblock In \emph{Proceedings of the IEEE/CVF conference on computer vision and pattern recognition}, 8238--8247.

\bibitem[{Perez et~al.(2018)Perez, Strub, De~Vries, Dumoulin, and Courville}]{perez2018film}
Perez, E.; Strub, F.; De~Vries, H.; Dumoulin, V.; and Courville, A. 2018.
\newblock Film: Visual reasoning with a general conditioning layer.
\newblock In \emph{Proceedings of the AAAI conference on artificial intelligence}, volume~32.

\bibitem[{Planamente et~al.(2022)Planamente, Plizzari, Alberti, and Caputo}]{planamente2022domain}
Planamente, M.; Plizzari, C.; Alberti, E.; and Caputo, B. 2022.
\newblock Domain generalization through audio-visual relative norm alignment in first person action recognition.
\newblock In \emph{Proceedings of the IEEE/CVF winter conference on applications of computer vision}, 1807--1818.

\bibitem[{Planamente et~al.(2024)Planamente, Plizzari, Peirone, Caputo, and Bottino}]{planamente2024relative}
Planamente, M.; Plizzari, C.; Peirone, S.~A.; Caputo, B.; and Bottino, A. 2024.
\newblock Relative norm alignment for tackling domain shift in deep multi-modal classification.
\newblock \emph{International Journal of Computer Vision}, 132(7): 2618--2638.

\bibitem[{Wang et~al.(2022)Wang, Lan, Liu, Ouyang, Qin, Lu, Chen, Zeng, and Philip}]{wang2022generalizing}
Wang, J.; Lan, C.; Liu, C.; Ouyang, Y.; Qin, T.; Lu, W.; Chen, Y.; Zeng, W.; and Philip, S.~Y. 2022.
\newblock Generalizing to unseen domains: A survey on domain generalization.
\newblock \emph{IEEE transactions on knowledge and data engineering}, 35(8): 8052--8072.

\bibitem[{Wang, Tran, and Feiszli(2020)}]{Wang_2020_CVPR}
Wang, W.; Tran, D.; and Feiszli, M. 2020.
\newblock What Makes Training Multi-Modal Classification Networks Hard?
\newblock In \emph{Proceedings of the IEEE/CVF Conference on Computer Vision and Pattern Recognition (CVPR)}.

\bibitem[{Wei et~al.(2024)Wei, Li, Feng, and Hu}]{wei2024diagnosing}
Wei, Y.; Li, S.; Feng, R.; and Hu, D. 2024.
\newblock Diagnosing and re-learning for balanced multimodal learning.
\newblock In \emph{European Conference on Computer Vision}, 71--86. Springer.

\bibitem[{Yu et~al.(2020)Yu, Kumar, Gupta, Levine, Hausman, and Finn}]{yu2020gradient}
Yu, T.; Kumar, S.; Gupta, A.; Levine, S.; Hausman, K.; and Finn, C. 2020.
\newblock Gradient surgery for multi-task learning.
\newblock \emph{Advances in neural information processing systems}, 33: 5824--5836.

\bibitem[{Zhou et~al.(2022)Zhou, Liu, Qiao, Xiang, and Loy}]{zhou2022domain}
Zhou, K.; Liu, Z.; Qiao, Y.; Xiang, T.; and Loy, C.~C. 2022.
\newblock Domain generalization: A survey.
\newblock \emph{IEEE Transactions on Pattern Analysis and Machine Intelligence}, 45(4): 4396--4415.

\end{thebibliography}

\end{document}